\documentclass[a4paper,11pt]{article}
\usepackage[utf8]{inputenc}

\usepackage{color}

\usepackage{amssymb}
\usepackage{amsmath}
\usepackage{amsfonts}
\usepackage{amsthm}
\usepackage{url}

\usepackage[dvips]{epsfig}
\usepackage[dvips]{graphics}

\usepackage{algorithm}
\usepackage{algorithmic}

\usepackage[all,knot,poly,arc]{xy}

\newcommand{\cX}{\mbox{$\cal X$}} 
\newcommand{\cY}{\mbox{$\cal Y$}}

\newcommand{\NN}{{\Bbb N}}
\newcommand{\RR}{{\Bbb R}}

\newcommand{\argmax}{\mbox{\rm argmax}}
\newcommand{\argmin}{\mbox{\rm argmin}}

\newcommand{{\uk}}{\mbox{$\underline{k}$}}

\def\nod(#1,#2){\put(#1,#2){\circle*{.125}}
\put(#1,#2){\makebox(0,0.5){{\small$#2$}}}}%
\def\rod(#1,#2){\put(#1,#2){\circle*{.2}}}
\def\NOD(#1,#2)#3{\put(#1,#2){\circle*{.2}}\put(#1,#2){\makebox(0,0.8){{\small$#3$}}}}

\def\EXX{{\hfill{$\diamondsuit$}}}

\newcounter{exampleNo}

\newtheorem{theorem}{Theorem}[section]

\newtheorem{proposition}[theorem]{Proposition}

\newenvironment{example}[1][Example \arabic{exampleNo}.]{\begin{trivlist}\refstepcounter{exampleNo}
\item[\hskip \labelsep {\bfseries #1}]}{\end{trivlist}}

\title{Inference in Graded Bayesian Networks}
\author{Robert Leppert, Karl-Heinz Zimmermann\footnote{Email: k.zimmermann@tuhh.de}\\
Hamburg University of Technology\\
21071 Hamburg, Germany}

\begin{document}
\maketitle
\begin{abstract}
Machine learning provides algorithms that can learn from data and make inferences or predictions on data.
Bayesian networks are a class of graphical models that allow to represent a collection of random variables and their condititional dependencies by directed acyclic graphs.
In this paper, an inference algorithm for the hidden random variables of a Bayesian network is given
by using the tropicalization of the marginal distribution of the observed variables.
By restricting the topological structure to graded networks,
an inference algorithm for graded Bayesian networks will be established
that evaluates the hidden random variables rank by rank and in this way yields the most probable states of the hidden variables.
This algorithm can be viewed as a generalized version of the Viterbi algorithm for graded Bayesian networks.
\end{abstract}
\medskip

\mbox{\bf AMS Subject Classification:} 62F15, 68T05, 16Y60
\medskip

\mbox{\bf Keywords:} Bayesian network, marginal distribution, rank function, tropical semiring, inference, Viterbi algorithm

\section{Introduction}
A Bayesian network is a statistical model which provides a graphical representation of probabilistic relationships between several random variables in the form of
a directed acyclic graph.
Such a network gives an efficient representation of the joint probability distribution by using the conditional dependencies between the variables.
These networks were introduced by Judea Pearl in the 1980s~\cite{pearl1} and attracted a great deal of attention in research and industry since the 1990s.
Today, Bayesian networks are widespread in artificial intelligence, knowledge engineering, and machine learning~\cite{barber,koski,pearl2,russell}.


In machine learning, statistical inference refers to the discovery of hidden states given observed states.
In Bayesian networks, statistical inference concerns the finding of the most probable states of the hidden variables given the observed variables.
Statistical inference can be utilized to answer probabilistic queries about hidden variables given instantiations of the observed variables, such as
diagnosis $P(\mbox{cause}|\mbox{symptom})$, 
prediction $P(\mbox{symptom}|\mbox{cause})$,
and classification $P(\mbox{class}|\mbox{data})$. 
However, this problem is NP-hard in the number of hidden variables~\cite{cooper}.

In practical terms, the approximation algorithms either make topological structural constraints such as in naive Bayesian networks~\cite{rish}
or restrictions on the conditional probabilities such as in the powerful bounded variance algorithm~\cite{dagum}.
The most popular approximation algorithms for calculating marginal distributions are based on sum-product message passing 
or belief propagation~\cite{barber,koski,pearl2,wiberg,yed}.
Other inference algorithms make use of importance sampling~\cite{doucet} or Markov chain Monte Carlo simulation~\cite{ritter}.

In this paper, an inference algorithm for the hidden random variables of a Bayesian network is given
by using the tropicalization of the marginal distribution of the observed variables.
By restricting the topological structure to graded networks,
an inference algorithm for graded Bayesian networks will be derived
which evaluates the hidden random variables rank by rank and in this way yields the most probable states of the hidden variables.

\section{Bayesian Networks} 
A Bayesian network is a probabilistic graphical model which represents a collection of random variables 
and their conditional dependencies in form of a directed acyclic graph (DAG).

Let $X_1,\ldots,X_n$ be random variables with state sets $\cX_1,\ldots,\cX_n$, respectively.
Then the random vector $X = (X_1,\ldots,X_n)$ has the state set $\cX = \cX_1\times \cdots\times \cX_n$.
By using conditional probabilities, the joint probability distribution of the random variables $X_1,\ldots,X_n$ factors as follows,
\begin{eqnarray}\label{e-p_X}
p_X = p_{X_1,\ldots,X_n} = \prod_{i=1}^n p_{X_i|X_{i+1},\ldots,X_n}.
\end{eqnarray}
If a DAG models the causal relationships between the random variables, factorizations of this kind can often be simplified
since some random variables may be conditionally independent of other ones.

For this, let $G=(V,E)$ be a DAG whose node set $V=\{v_1,\ldots,v_n\}$ corresponds one-to-one with a collection of the random variables $X_1,\ldots,X_n$.
Write $\Pi(X_i)$ for the parent set of the random variable $X_i$ in the DAG, $1\leq i\leq n$.
A topological sorting of a directed graph is a linear ordering $\leq$ of the vertices in which the vertices of every directed edge $u\rightarrow v$ 
are ordered such that $u$ comes before $v$ in the linear ordering.
Such an ordering is possible if and only if the graph is a DAG.
Thus by topological sorting, there is an ordering $(v_{\sigma(1)},\ldots, v_{\sigma(n)})$ of the nodes 
such that for each $1\leq i\leq n$, the parent set of the node $v_{\sigma(i)}$ is a subset of
$\{v_{\sigma(1)},\ldots, v_{\sigma(i-1)}\}$. 
For a finite node set $V$, the vertices of a DAG $G=(V,E)$ can be sorted topologically in $O(|V|+|E|)$ steps~\cite{kahn}.

A {\em Bayesian network\/} is a pair $(G,p)$
consisting of a DAG $G=(V,E)$ with node set $V=\{v_1,\ldots,v_n\}$ for some integer $n\geq 1$,
which corresponds one-to-one with a collection of random variables $X_1,\ldots,X_n$,
and a collection of conditional probability distributions $p$ of the random variables 
such that the following holds:
\begin{itemize}
\item
For each node $v_i\in V$, which has no parent,
there is a probability distribution $p_{X_i}$ of the random variable $X_i$.
\item
For each node $v_i\in V$, which has a non-empty parent set $\Pi(X_i)$,
there is a conditional probability distribution $p_{X_i|\Pi(X_i)}$.
\item
The joint probability function $p_{X_1,\ldots,X_n}$ factors using the conditional probability distribution functions $p_{X_i|\Pi(X_i)}$ as follows,
\begin{eqnarray}\label{e-p-joint}
p_{X_1,\ldots,X_n} = \prod_{i=1}^n p_{X_i|\Pi(X_i)}.
\end{eqnarray}
\end{itemize}
The shape of factorization follows the {\em Markov property\/}\index{Markov property} which states that each random variable depends directly only on its parents.
\begin{example}
Consider the Bayesian network with the random variables $X_1,\ldots,X_4$ in Fig.~\ref{f-BN-4}.
The parent sets are
$\Pi(X_1) = \emptyset$,
$\Pi(X_2) = \{X_1\}$,
$\Pi(X_3) = \{X_1\}$,
and
$\Pi(X_4) = \{X_2,X_3\}$.
The joint probability function factors as follows,
\begin{eqnarray}
p_{X_1,X_2,X_3,X_4} &=& p_{X_1} p_{X_2|X_1} p_{X_3|X_2,X_1} p_{X_4|X_3,X_2,X_1}\\
&=&  p_{X_1} p_{X_2|X_1} p_{X_3|X_1} p_{X_4|X_3,X_2}.\nonumber
\end{eqnarray}
\EXX 
\end{example}
\begin{figure}[hbt]
\begin{center}
\mbox{$
\xymatrix{
                             & *++[o][F-]{X_1} \ar@{->}[dr] \ar@{->}[dl] & \\
*++[o][F-]{X_2} \ar@{->}[dr] &                                         &*++[o][F-]{X_3} \ar@{->}[dl] \\
                             & *++[o][F-]{X_4}                           & \\
}$}
\end{center}
\caption{A DAG.}\label{f-BN-4}
\end{figure}

\begin{example}
A Bayesian network for printer troubleshooting adapted from the operating system Microsoft Windows~95 has 24 variables as shown in Fig.~\ref{f-BN-pts}~\cite{heck}.
Suppose all random variables have binary state sets.
Then the joint probability distribution has $2^{24}=16,777,216$ entries.
However, as a Bayesian network, the number of conditional distributions that need to be specified is only
$15\cdot 1 + 0\cdot 2^1+4\cdot 2^2 + 2\cdot 2^3 + 3\cdot 2^4 = 95$.
\EXX
\end{example}
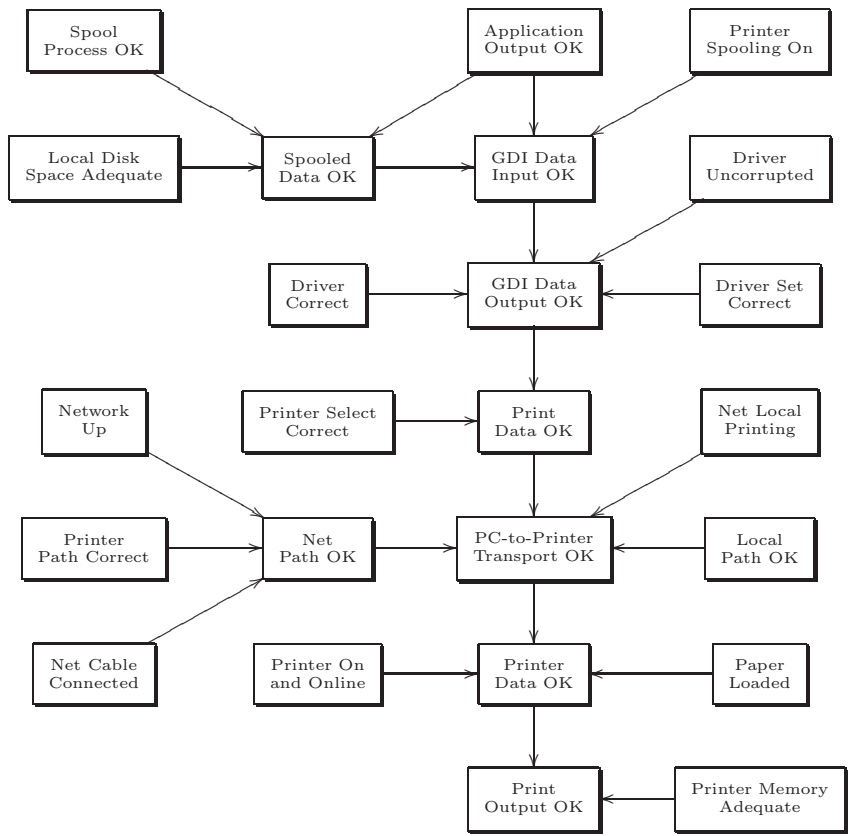
\begin{figure}[hbt]
\begin{center}
\tiny
\mbox{$
\xymatrix{
*++[F-,]{\txt{Spool\\Process OK}}\ar@{->}[rd] &                         &  *++[F-,]{\txt{Application\\Output OK}}\ar@{->}[d]\ar@{->}[ld] &  *++[F-,]{\txt{Printer\\Spooling On}}\ar@{->}[dl] & \\
*++[F-,]{\txt{Local Disk\\Space Adequate}}\ar@{->}[r]&  *++[F-,]{\txt{Spooled\\Data OK}}\ar@{->}[r] & *++[F-,]{\txt{GDI Data\\Input OK}}\ar@{->}[d] &  *++[F-,]{\txt{Driver\\Uncorrupted}}\ar@{->}[dl] & \\
                                    & *++[F-,]{\txt{Driver\\Correct}}\ar@{->}[r] & *++[F-,]{\txt{GDI Data\\Output OK}}\ar@{->}[d] & *++[F-,]{\txt{Driver Set\\Correct}}\ar@{->}[l] &\\ 
*++[F-,]{\txt{Network\\Up}}\ar@{->}[rd] & *++[F-,]{\txt{Printer Select\\Correct}}\ar@{->}[r] & *++[F-,]{\txt{Print\\Data OK}}\ar@{->}[d] & *++[F-,]{\txt{Net Local\\Printing}}\ar@{->}[dl] \\ 
*++[F-,]{\txt{Printer\\Path Correct}}\ar@{->}[r]  & *++[F-,]{\txt{Net\\Path OK}}\ar@{->}[r]&  *++[F-,]{\txt{PC-to-Printer\\Transport OK}}\ar@{->}[d] & *++[F-,]{\txt{Local\\Path OK}}\ar@{->}[l]  \\ 
*++[F-,]{\txt{Net Cable\\Connected}}\ar@{->}[ur]  & *++[F-,]{\txt{Printer On\\and Online}}\ar@{->}[r] & *++[F-,]{\txt{Printer\\Data OK}}\ar@{->}[d] & *++[F-,]{\txt{Paper\\Loaded}}\ar@{->}[l]&\\
                                   &                                                   & *++[F-,]{\txt{Print\\Output OK}} & *++[F-,]{\txt{Printer Memory\\Adequate}}\ar@{->}[l] & \\
}
$}
\end{center}
\caption{A Bayesian network for printer troubleshooting.}\label{f-BN-pts}
\end{figure}

\section{Inference Algorithm}
In this section, we present an inference algorithm for the hidden random variables of a Bayesian network
by using the tropicalization of the marginal distribution of the observed variables.
By restricting the topological structure to graded networks,
an inference algorithm for graded Bayesian networks will be obtained
that evaluates the hidden random variables rank by rank and in this way yields the most probable states of the hidden variables.
This algorithm can be viewed as a generalized version of the Viterbi algorithm for graded Bayesian networks.

For this, let $(G,p)$ be a Bayesian network given by the DAG $G=(V,E)$ and the global probability distribution $p$ as defined in~(\ref{e-p-joint}).
The node set $V=\{v_1,\ldots,v_{m+n}\}$ with $m,n\geq 1$ is assumed to correspond one-to-one 
with a collection of $m+n$ random variables denoted by $X_1,\ldots,X_m,Y_1,\ldots,Y_n$.
Suppose the variables $X_1,\ldots,X_m$ are observed or instantiated and the variables $Y_1,\ldots,Y_n$ are unobserved or hidden. 
We may assume that the collection of random variables is sorted topologically such that 
$X_1> \ldots >X_m$ and $Y_1>\ldots> Y_n$.
Note that according to this sorting, the variable $X_1$ has only parents in the hidden variables and the variable $Y_1$ has only parents in the observed variables.
Let the variables $X_1,\ldots,X_m$ and $Y_1,\ldots,Y_n$ have finite state sets $\cX_1,\ldots,\cX_m$ and $\cY_1,\ldots,\cY_n$, respectively.
Then the random vectors $X=(X_1,\ldots,X_m)$ and $Y=(Y_1,\ldots,Y_n)$ have state sets 
$\cX=\cX_1\times\ldots\times\cX_m$ and $\cY=\cY_1\times\ldots\times\cY_n$, respectively.
Thus the global probability distribution factors as follows,
\begin{eqnarray}
p_{X,Y} = \prod_{i=1}^m p_{X_i|\Pi(X_i)} \prod_{j=1}^n p_{Y_j|\Pi(Y_j)}. 
\end{eqnarray}
The probability of the observed sequence $x\in\cX$ is given by the marginal distribution
\begin{eqnarray}\label{e-md}
p_{X} (x) = \sum_{y\in{\cal Y}} p_{X,Y}(x,y).
\end{eqnarray}


The variables in the DAG $D=(V,E)$ can be equipped with a {\em semi-ranking function\/} $\rho$ from $V$ to $\NN_0$.
For this, each variable $Z$ with empty parent set or parent set in the observed variables is given the semi-rank $\rho(Z)=0$.
Since the graph is a DAG, there is at least one variable with semi-rank~0.
Moreover, each hidden variable $Z$ whose parents have already assigned semi-ranks is given the semi-rank 
\begin{eqnarray}
\rho(Z) = \max\{\rho(U)\mid \mbox{$U$ hidden and parent of $Z$}\}+1.
\end{eqnarray}
Furthermore, each observed variable $Z$ is given the largest semi-rank of its hidden parents,
\begin{eqnarray}
\rho(Z) = \max\{\rho(U)\mid \mbox{$U$ hidden and parent of $Z$}\}.
\end{eqnarray}
The reason is that the conditional probability $p_{Z|\Pi(Z)}$ of observed variable $Z$ with given value $Z=z$ can be evaluated as soon as the parents 
are instantiated (Ex.~\ref{e-4}).

Let $\rho_{\max}$ denote the maximal semi-rank of the nodes in the DAG $G$ 
and let $X_1^{(r)},\ldots,X_{s_r}^{(r)}$ and $Y_1^{(r)},\ldots,Y_{t_r}^{(r)}$ denote the collections of observed and hidden random variables with semi-rank $r$, 
$0\leq r\leq \rho_{\max}$, respectively. 
Then we have $\sum_{r=0}^{\rho_{\max}} s_r = m$ and $\sum_{r=0}^{\rho_{\max}} t_r = n$.
Moreover, the state set of the hidden variables with semi-rank $r$ is denoted by
\begin{eqnarray}
D(r) = \cY_1^{(r)}\times \ldots\times \cY_{t_r}^{(r)},\quad 0\leq r\leq \rho_{\max}.
\end{eqnarray}
Then by the semi-ranks of the nodes, the marginal distribution~(\ref{e-md}) can be written according to the following sum-product decomposition,
\begin{eqnarray}\label{e-sp}
p_{X}(x) 
&=& \left(\sum_{y\in D(0)} \prod_{i=1}^{s_0}p_{X_i^{(0)}|\Pi(X_i^{(0)})} \prod_{j=1}^{t_0}p_{Y_j^{(0)}|\Pi(Y_j^{(0)})}\right.\nonumber\\
&& \cdot \left(\sum_{y\in D(1)} \prod_{i=1}^{s_1}p_{X_i^{(1)}|\Pi(X_i^{(1)})} \prod_{j=1}^{t_1}p_{Y_j^{(1)}|\Pi(Y_j^{(1)})}\right.\nonumber\\
&&\ldots\\
&& \cdot \left(\left.\sum_{y\in D(\rho)} \prod_{i=1}^{s_\rho}p_{X_i^{(\rho)}|\Pi(X_i^{(\rho)})} \prod_{j=1}^{t_\rho}p_{Y_j^{(\rho)}|\Pi(Y_j^{(\rho)})}\right)\ldots\right), \nonumber
\end{eqnarray}
where $\rho=\rho_{\max}$ and the arguments of the conditional probabilities have been omitted for readability.
This decomposition is sound, since the computation in the $r$-th bracket 
corresponding to the collections of variables $X_i^{(r)}$ and $Y_j^{(r)}$ of semi-rank $r$ depends on the parent nodes which are of lower semi-rank.


\begin{example}\label{e-5}
Consider the Bayesian network given by the DAG in Fig.~\ref{f-N5}.
Take the topological sorting $X_1>Y_1>Y_2>Y_3>Y_4>Y_5$.
The random variables have semi-ranks $\rho(X_1) = \rho(Y_1)=0$, $\rho(Y_2)=\rho(Y_3)=1$, $\rho(Y_4)=2$, and $\rho(Y_5)=3$.
In view of the DAG, the joint probability distribution factors as follows,
\begin{eqnarray}
p_{X,Y} = p_{X_1}p_{Y_1|X_1}p_{Y_2|Y_1}p_{Y_3|Y_1}p_{Y_4|Y_2}p_{Y_5|Y_3,Y_4}.
\end{eqnarray}
The marginal distribution of the observed value $x_1\in\cX_1$ can be written as follows,
\begin{eqnarray}
p_{X_1}(x_1) &=& \sum_{(y_1,\ldots,y_5)\in {\cal Y}} p_{X,Y}(x_1,y_1,\ldots,y_5)\nonumber\\
&=& p_{X_1}(x_1)\cdot \left( \sum_{y_1\in{\cal Y}_1} p_{Y_1|X_1}(y_1|x_1) \right.\nonumber\\
&& \cdot \left( \sum_{(y_2,y_3)\in{\cal Y}_2\times {\cal Y}_3} p_{Y_2|Y_1}(y_2|y_1)p_{Y_3|Y_1}(y_3|y_1) \right.\\
&& \cdot \left( \sum_{y_4\in{\cal Y}_4} p_{Y_4|Y_2}(y_4|y_2)\right.\nonumber\\
&& \cdot \left( \left.\sum_{y_5\in{\cal Y}_5} p_{Y_5|Y_3,Y_4}(y_5|y_3,y_4)\right)\ldots\right). \nonumber
\end{eqnarray}
\EXX
\end{example}
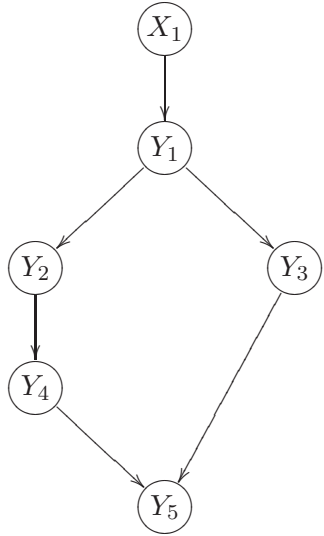
\begin{figure}[hbt]
\begin{center}
\mbox{$
\xymatrix{
                           & *++[o][F-]{X_1}\ar@{->}[d] & \\
                           & *++[o][F-]{Y_1}\ar@{->}[rd]\ar@{->}[ld] & \\
*++[o][F-]{Y_2}\ar@{->}[d] &                                         & *++[o][F-]{Y_3}\ar@{->}[ddl] \\
*++[o][F-]{Y_4}\ar@{->}[dr] & \\
& *++[o][F-]{Y_5} \\
}
$}
\end{center}
\caption{A non-graded Bayesian network.}\label{f-N5}
\end{figure}

\begin{example}\label{e-4}
Consider the Bayesian network given by the DAG in Fig.~\ref{f-bn1}.
We have $X_1>X_2>X_3$ and $Y_1>Y_2>Y_3>Y_4$, and the random variables have the semi-ranks
$\rho(X_1)=\rho(X_2)=\rho(Y_1)=0$, $\rho(X_3)=\rho(Y_2)=r(Y_3)=1$, and $\rho(Y_4)=2$.
In view of the DAG, the joint probability distribution factors as follows,
\begin{eqnarray}
p_{X,Y} = p_{X_1}p_{X_2|X_1}p_{Y_1|X_1}p_{Y_2|X_2,Y_1}p_{Y_3|Y_1}p_{X_3|Y_2}p_{Y_4|Y_2,Y_3}.
\end{eqnarray}
The marginal distribution of the observed sequence $(x_1,x_2,x_3)\in\cX$ can be decomposed as follows, 
\begin{eqnarray}
\lefteqn{p_{X}(x_1,x_2,x_3)} \nonumber\\
&=& \sum_{(y_1,y_2,y_3,y_4)\in{\cal Y}} p_{X,Y} (x_1,x_2,x_3,y_1,y_2,y_3,y_4) \\
&=& p_{X_1}(x_1)p_{X_2|X_1}(x_2|x_1) \cdot \left(\sum_{y_1\in{\cal Y}_1} p_{Y_1|X_1}(y_1|x_1)\right. \nonumber\\
&& \cdot \left(\sum_{(y_2,y_3)\in{\cal Y}_2\times {\cal Y}_3}p_{Y_2|X_2,Y_1}(y_2|x_2,y_1) p_{Y_3|Y_1}(y_3|y_1)p_{X_3|Y_2}(x_3|y_2)\right. \nonumber \\
&& \cdot \left(\left.\sum_{y_4\in{\cal Y}_4} p_{Y_4|Y_2,Y_3}(y_4|y_2,y_3)\right)\ldots\right)\nonumber.
\end{eqnarray}
\EXX
\end{example}
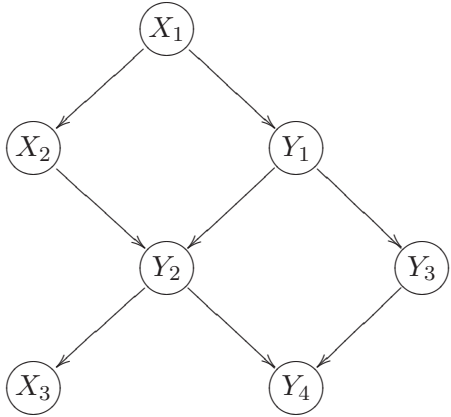
\begin{figure}[hbt]
\begin{center}
\mbox{$
\xymatrix{
                           & *++[o][F-]{X_1}\ar@{->}[dr]\ar@{->}[dl] &   \\                         
*++[o][F-]{X_2}\ar@{->}[dr]&                                         & *++[o][F-]{Y_1}\ar@{->}[dl]\ar@{->}[dr] & \\
                           & *++[o][F-]{Y_2}\ar@{->}[dr]\ar@{->}[dl] &                                         & *++[o][F-]{Y_3}\ar@{->}[dl]\\
*++[o][F-]{X_3}            &                                         &  *++[o][F-]{Y_4} \\
}
$}
\end{center}
\caption{A graded Bayesian network.}\label{f-bn1}
\end{figure}


The marginal distribution of the observed random variables can be used for the probabilistic inference of the hidden random variables,
which amounts to finding the most probable state sequences of the hidden variables.
This can be achieved by tropicalization of the marginal distribution of the observed variables.
For this, we introduce the tropical semiring~\cite{sturm}.

A {\em semiring\/} is an algebraic structure similar to a ring, 
but without the requirement that each element must have an additive inverse.
More specifically, a semiring is a non-empty set $R$ together with two binary operations, called addition $+$ and multipli\-cation~$\cdot$, such that
$(R,+)$ is a commutative monoid with identity element~0,
$(R,\cdot)$ is a monoid with identity element~1,
the multiplication distributes over addition, i.e., for all $a,b,c\in R$,
\begin{eqnarray}
a\cdot(b+c) = (a\cdot b)+(a\cdot c)\quad\mbox{and}\quad (a+b)\cdot c = (a\cdot c) + (b\cdot c),
\end{eqnarray}
and the multiplication with~0 annihilates $R$, i.e., for all $a\in R$, $a\cdot 0 = 0 = 0\cdot a$.

A semiring is {\em commutative\/} if its multiplication is commutative, i.e., for all $a,b\in R$, $a\cdot b = b\cdot a$.
A semiring is {\em idempotent\/} if its addition is idempotent, i.e., for all $a\in R$, $a+a=a$.

For instance, each ring is also a semiring.
Moreover, the set of natural numbers $\NN_0$ forms a commutative semiring with the ordinary addition and multiplication.
Likewise, the set of non-negative real numbers forms a commutative semiring.

The set $\RR\cup\{\infty\}$ together with the operations
\begin{eqnarray}
x\oplus y = \min\{x,y\}\quad\mbox{and}\quad x\odot y =x+y,\quad x,y\in\RR\cup\{\infty\},
\end{eqnarray}
with $x+\infty=\infty$ for all $x\in\RR\cup\{\infty\}$
forms an idempotent commutative semiring with additive identity $\infty$ and multiplicative identity~0.
Note that additive and multiplicative inverses may not exist in a semiring. 
For instance, the equations $1\oplus x = 2$ and $\infty \odot x = 1$ have no solutions $x\in \RR\cup\{\infty\}$.
This semiring is known as {\em tropical semiring}.
The attribute "tropical" was coined by French scholars (1998) in honor of the Brazilian mathematician Imre Simon who studied the tropical semiring in the early 1960s.

The mapping $\phi:\RR_{\geq 0}\rightarrow\RR\cup \{\infty\}:x\mapsto -\log x$ is bijective and monotonically decreasing with 
$\phi(0)=\infty$, $\phi(1)=0$, and
\begin{eqnarray}
\phi(x\cdot y) = \phi(x)\odot \phi(y),\quad x,y\in\RR_{\geq 0}.
\end{eqnarray}
The mapping $\phi$ is the {\em tropicalization\/} of the ordinary semiring $(\RR_{\geq 0},+,\cdot)$. 
In this way, large values (probabilities) are mapped to small values (weights) and vice versa.

Given an observed sequence $x\in \cX$, the objective is to find one (or all) sequences $y\in \cY$ with maximum likelihood
\begin{eqnarray}
p_{Y|X} (y\mid x) = \frac{p_{X,Y}(x,y)}{p_{X}(x)}.
\end{eqnarray}
Since the observed sequence $x$ is fixed, the likelihood $p_{Y|X} (y\mid x)$  is directly proportional to the joint probability
$p_{X,Y}(x,y)$ provided that $p_{X}(x)>0$.
Suppose that $p_{X}(x)>0$.
Then the aim is to find one (or all) sequences $\bar y\in \cY$ with the property
\begin{eqnarray}
\bar y    &=& \argmax_{y\in {\cal Y}} \{ p_{X,Y}(x,y)\}.
\end{eqnarray}
Each optimal sequence $\bar y$ is called an {\em explanation\/} of the given sequence $x$. 
The explanations can be found by tropicalization.
For this, put $w_{X,Y}(x,y) = -\log p_{X,Y}(x,y)$ and $w_X(x) = -\log p_{X}(x)$ for all $x\in\cX$ and $y\in \cY$. 
Then the tropicalization yields 
\begin{eqnarray}
w_X(x) = \bigoplus_{y\in{\cal Y}} w_{X,Y}(x,y).
\end{eqnarray}
The explanations $\bar y$ can be obtained by evaluation in the tropical semiring, 
\begin{eqnarray}
\bar y   &=& \argmin_{y\in {\cal Y}} \{w_{X,Y}(x,y)\}.
\end{eqnarray}

The value $w_X(x)$ can be computed by tropicalizing the sum-product decomposition of the marginal probability $p_{X}(x)$.
For this, we put $w_{Z|\Pi(Z)} = -\log p_{Z|\Pi(Z)}$ for each random variable $Z$.
Thus if in the sum-product decomposition~(\ref{e-sp}) sums are replaced by tropical sums and products by tropical products, we obtain 
\begin{eqnarray}\label{e-spw}
w_{X}(x) 
&=& \left(\bigoplus_{y\in D(0)} \bigodot_{i=1}^{s_0}w_{X_i^{(0)}|\Pi(X_i^{(0)})} \odot \bigodot_{j=1}^{t_0}w_{Y_j^{(0)}|\Pi(Y_j^{(0)})}\right.\nonumber\\
&& \odot \left(\bigoplus_{y\in D(1)} \bigodot_{i=1}^{s_1}w_{X_i^{(1)}|\Pi(X_i^{(1)})} \odot \bigodot_{j=1}^{t_1}w_{Y_j^{(1)}|\Pi(Y_j^{(1)})}\right.\nonumber\\
&&\ldots\\
&& \odot \left(\left.\bigoplus_{y\in D(\rho)} \bigodot_{i=1}^{s_\rho}w_{X_i^{(\rho)}|\Pi(X_i^{(\rho)})} \odot \bigodot_{j=1}^{t_\rho}w_{Y_j^{(\rho)}|\Pi(Y_j^{(\rho)})}\right)\ldots\right), \nonumber
\end{eqnarray}
where $\rho=\rho_{\max}$ and the arguments of the weights have been omitted for readability.
This yields the following result.
\begin{proposition}
Let $x\in\cX$.
The tropicalization $w_X(x)$ of the marginal probability $p_{X}(x)$ provides the explanations of the sequence $x$.
\end{proposition}

However, the tropicalization of the marginal probability does not overcome the NP-hardness of the inference problem.
Our aim is to provide an easily structured inference algorithm for a class of topologically constrained Bayesian networks which emerge quite naturally in practice.

For this, a DAG $G=(V,E)$  is called {\em graded\/} if it can be equipped with a {\em rank function\/} $\rho$ from $V$ to $\NN_0$.
A rank function of a DAG must be compatible with the given topological ordering and the rank must be consistent with the covering relation of the ordering~\cite{stanley}.
In our case, each variable $Z$ with empty parent set or parent set in the observed variables is given the rank $\rho(Z)=0$.
Moreover, each hidden variable\/ $Z$ is assigned the rank $\rho=\rho(Z)\geq 1$ if all its hidden parents have rank $\rho-1$.
Furthermore, each observed variable\/ $Z$ is assigned the rank $\rho=\rho(Z)$ if all its hidden parents have rank $\rho$.
For instance, the DAG in Fig.~\ref{f-bn1} is graded, while the DAG in Fig.~\ref{f-N5} is not.
A Bayesian network is {\em graded\/} if its underlying DAG is graded.

Inference in a graded Bayesian network has the advantage that in the computation of the $r$-th expression
\begin{eqnarray}\label{e-grad}
\bigoplus_{y\in D(r)} 
\bigodot_{i=1}^{s_r}w_{X_i^{(r)}|\Pi(X_i^{(r)})}
\odot 
\bigodot_{j=1}^{t_r}w_{Y_j^{(r)}|\Pi(Y_j^{(r)})},\quad 1\leq r\leq \rho_{\max},
\end{eqnarray}
the terms  
$w_{X_i^{(r)}|\Pi(X_i^{(r)})}$ and $w_{Y_j^{(r)}|\Pi(Y_j^{(r)})}$
depend only on the parent values $y'\in D(r-1)$ of the hidden variables of previous rank $r-1$.
In this way, the evalution of the expression $w_X(x)$ has a simple bookkeeping structure (Alg.~\ref{a-inf}).
By the gradedness of the nodes, the hidden parents of each hidden variable $Y$ with rank $r=\rho(Y)\geq 1$ all have rank $r-1$ 
and so the computation of array element $A[r,y]$ with rank $r$ and $y\in D(r)$ requires only the array elements $A[r-1,y']$ with $y'\in D(r-1)$ of the previous rank.

The algorithm follows the principle of dynamic programming~\cite{bellman} 
and consists of a forward algorithm evaluating the tropicalized expression $w_X(x)$ and a backward algorithm 
which provides one or all explanations of the collection of hidden variables.
The latter is achieved by recording in each step $r$ all state values in $D(r)$ which attain the minimum in the minimization step,  $0\leq r\leq \rho_{\max}$. 
This information can already be recorded by the forward algorithm.
Then the trace back of all optimal decisions made in each step can provide all explanations.
The forward algorithm evaluates the expression~(\ref{e-spw}) by using an array $A$ 
such that the array entries $A[r,y]$ with $y\in D(r)$ record all decisions made up to the variables of rank $r$. 
\begin{algorithm}
\caption{Forward inference algorithm.}\label{a-inf}
\begin{algorithmic}
\REQUIRE Graded Bayesian network $(G,p)$,
observed sequence $x\in\cX$, family of scores $w_{X_i|\Pi(X_i)}$ and $w_{Y_j|\Pi(Y_j)}$
\ENSURE Score $w_X(x)$
\STATE $A \leftarrow {\rm array}[r,|D(r)|]_{r=0}^{\rho_{\max}}$ 
\COMMENT{array $A$ has varying column size}
\FOR{$y \in D(0)$}
\STATE $A[0,y] \leftarrow \sum_{i=1}^{s_0} w_{X_i^{(0)}} + \sum_{i=1}^{t_0} w_{Y_i^{(0)}}(y)$
\ENDFOR
\FOR{$r\leftarrow 1$ to $\rho_{\max}$}
\FOR{$y \in D(r)$}
\STATE $A[r,y] \leftarrow \min_{y'\in D(r-1)} \left\{A[r-1,y'] + \sum_{i=1}^{s_r} w_{X_i^{(r)}|\Pi(X_i^{(r)})} \right.$
\STATE \hfill $\left. + \sum_{j=1}^{t_r} w_{Y_j^{(r)}|\Pi(Y_j^{(r)})}\right\}$
\ENDFOR
\ENDFOR
\STATE $w\leftarrow \min_{y\in R(\rho_{\max})} \left\{ A[\rho_{\max},y]\right\}$.
\RETURN $w$
\end{algorithmic}
\end{algorithm}

The complexity of the evaluation of the tropicalized term $w_X(x)$ depends on the underlying DAG.
The array $A$ has size $\sum_{r=0}^{\rho_{\max}} |D(r)|$ and the computation of array element $A[r,y]$ requires $O(|D(r-1)|\cdot (s_r+t_r))$ steps.
Suppose all state sets have $l$ elements.
Then we have $D(r)=l^{t_r}$ for all $0\leq r\leq \rho_{\max}$.

In the best case, the hidden random variables all have the same rank and common observed ascendants.
Then $\rho_{\max}=0$.
In view of the graded DAG in Fig.~\ref{f-rank0}, the random variables have ranks $\rho(X_1)=\rho(Y_1)=\ldots=\rho(Y_n) = 0$.
Since the minimization is decoupled, the inference algorithm has time complexity $O(ln)$ and computes for each observed value $x_1\in\cX_1$ the following,
\begin{eqnarray}
w_{X_1}(x_1) 
&=& \min_{y_1,\ldots,y_n} \left(w_{Y_1|X_1}(y_1|x_1) + \ldots + w_{Y_n|X_1}(y_n|x_1)\right)\\
&=& \min_{y_1} \left(w_{Y_1|X_1}(y_1|x_1)\right) + \ldots + \min_{y_n} \left(w_{Y_n|X_1}(y_n|x_1)\right).\nonumber
\end{eqnarray}
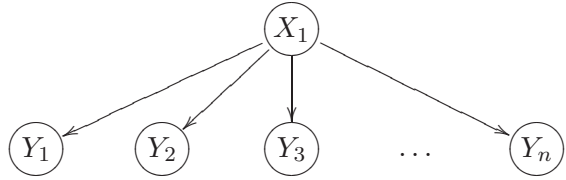
\begin{figure}[hbt]
\begin{center}
\mbox{$
\xymatrix{
                &                 & *++[o][F-]{X_1}\ar@{->}[dll]\ar@{->}[dl]\ar@{->}[d]\ar@{->}[drr] \\
*++[o][F-]{Y_1} & *++[o][F-]{Y_2} & *++[o][F-]{Y_3}                                & \ldots & *++[o][F-]{Y_n} \\
}
$}
\end{center}
\caption{A graded Bayesian network.}\label{f-rank0}
\end{figure}

\begin{figure}[hbt]
\begin{center}
\mbox{$
\xymatrix{
*++[o][F-]{Y_1}\ar@{->}[r]\ar@{->}[d] &  *++[o][F-]{Y_2}\ar@{->}[r]\ar@{->}[d] & \ldots\ar@{->}[r] & *++[o][F-]{Y_{n-1}}\ar@{->}[r]\ar@{->}[d] & *++[o][F-]{Y_n}\ar@{->}[d] \\
*++[o][F-]{X_1}                       & *++[o][F-]{X_2}                        & \ldots &  *++[o][F-]{X_{n-1}} & *++[o][F-]{X_n} \\
}
$}
\end{center}
\caption{Hidden Markov model.}\label{f-hmm}
\end{figure}
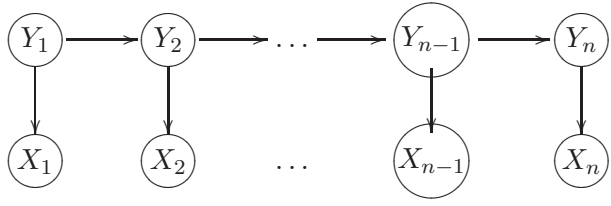

In the hidden Markov model (HMM) the hidden random variables form a chain.
Here $\rho_{\max}=n-1$ and $t_r=1$ for each $0\leq r\leq \rho_{\max}$.
In view of the graded DAG in Fig.~\ref{f-hmm}, the random variables have ranks $\rho(X_r)=\rho(Y_r)=r-1$ for $1\leq r\leq n$.
The Viterbi algorithm~\cite{rab,sturm,zim} calculates for each observed sequence $x=x_1\ldots x_n\in\cX$ the following,
\begin{eqnarray}
A[0,y] &:=& w_{X_1}(x_1) + w_{Y_1}(y),\nonumber\\
A[1,y] &:=& \min_{y_1} \left(w_{Y_2|Y_1} (y|y_1) + w_{X_2|Y_2} (x_2|y) + A[0,y_1]\right)\nonumber\\
&& \ldots\\
A[n-1,y] &:=& \min_{y_{n-1}} \left(w_{Y_n|Y_{n-1}} (y|y_{n-1}) + w_{X_n|Y_n} (x_n|y) + A[n-2,y_{n-1}]\right)\nonumber\\
w_X(x) &:=& \min_{y_n} A[n-1,y_n]\nonumber.
\end{eqnarray}
The array has size $n\cdot l$ and the computation of each array element requires $O(l)$ steps.
Hence, the time complexity is $O(l^2n)$. 
Note that the Bayesian networks for 
the hidden tree Markov model~\cite{fels} and
stochastic automata~\cite{zimstoch} 
are graded as well and their inference algorithms have both the same time complexity $O(l^2n)$.

\begin{figure}[hbt]
\begin{center}
\mbox{$
\xymatrix{
*++[o][F-]{Y_1}\ar@{->}[drr] & *++[o][F-]{Y_2}\ar@{->}[dr] & *++[o][F-]{Y_3}\ar@{->}[d] & \ldots & *++[o][F-]{Y_n}\ar@{->}[dll] \\
                             &                             & *++[o][F-]{X_1}            &        & \\
}
$}
\end{center}
\caption{A graded Bayesian network.}\label{f-rank00}
\end{figure}
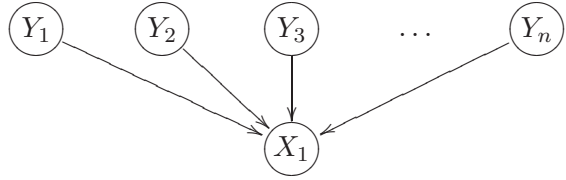

In the worst case, the hidden random variables have the same rank and common observed descendants.
Then $\rho_{\max} = 0$.
In view of the graded DAG in Fig.~\ref{f-rank00}, the random variables have ranks $\rho(Y_1)=\ldots=\rho(Y_n)=\rho(X_1)=0$.
Since the minimization is fully coupled, the inference algorithm has time complexity $O(l^n n)$ and computes for each observed value $x_1\in\cX_1$ the following,
\begin{eqnarray}
w_{X_1}(x_1) = \min_{y_1,\ldots,y_n} \left( w_{X_1|Y_1,\ldots,Y_n} (x_1|y_1,\ldots,y_n) +\sum_{i=1}^n w_{Y_i}(y_i)\right).
\end{eqnarray}




\begin{example}
In view of the Bayesian network in Ex.~\ref{e-4}, the tropicalization of the marginal distribution $p_X(x_1,x_2,x_3)$ gives
\begin{eqnarray}
\lefteqn{w_{X}(x_1,x_2,x_3)}\nonumber\\ 
&=& \bigoplus_{(y_1,y_2,y_3,y_4)\in{\cal Y}} w_{X,Y} (x_1,x_2,x_3,y_1,y_2,y_3,y_4) \\
&=& w_{X_1}(x_1)\odot w_{X_2|X_1}(x_2|x_1) \odot \left(\bigoplus_{y_1\in{\cal Y}_1} w_{Y_1|X_1}(y_1|x_1)\right. \nonumber\\
&& \odot \left(\bigoplus_{(y_2,y_3)\in{\cal Y}_2\times {\cal Y}_3}w_{Y_2|X_2,Y_1}(y_2|x_2,y_1) \odot w_{Y_3|Y_1}(y_3|y_1)\odot w_{X_3|Y_2}(x_3|y_2)\right. \nonumber\\
&& \odot \left(\left.\bigoplus_{y_4\in{\cal Y}_4} w_{Y_4|Y_2,Y_3}(y_4|y_2,y_3)\right)\ldots\right)\nonumber.
\end{eqnarray}

Assume that the hidden variables $Y_1,\ldots,Y_4$ have common state set $\Sigma=\{a,b\}$.
Then we have $D(0)=\Sigma$, $D(1)=\Sigma^2$, and $D(2)=\Sigma$.
The forward inference algorithm computes the following:

{\small
\begin{eqnarray*}
A[0,a] &=& w_{Y_1|X_1}(a|x_1) + w_{X_1}(x_1) + w_{X_2|X_1}(x_2|x_1),\\ 
A[0,b] &=& w_{Y_1|X_1}(b|x_1) + w_{X_1}(x_1) + w_{X_2|X_1}(x_2|x_1),\\ 
A[1,aa] &=& \min_{y_1\in D(0)}\left(A[0,y_1] + w_{Y_2|X_2,Y_1}(a|x_2,y_1) + w_{Y_3|Y_1}(a|y_1) + w_{X_3|Y_2}(x_3|a)\right),\\
A[1,ab] &=& \min_{y_1\in D(0)}\left(A[0,y_1] + w_{Y_2|X_2,Y_1}(a|x_2,y_1) + w_{Y_3|Y_1}(b|y_1) + w_{X_3|Y_2}(x_3|a)\right),\\
A[1,ba] &=& \min_{y_1\in D(0)}\left(A[0,y_1] + w_{Y_2|X_2,Y_1}(b|x_2,y_1) + w_{Y_3|Y_1}(a|y_1) + w_{X_3|Y_2}(x_3|b)\right),\\
A[1,bb] &=& \min_{y_1\in D(0)}\left(A[0,y_1] + w_{Y_2|X_2,Y_1}(b|x_2,y_1) + w_{Y_3|Y_1}(b|y_1) + w_{X_3|Y_2}(x_3|b)\right),\\
A[2,a] &=& \min_{y_2y_3\in D(1)}\left(A[1,y_2y_3] + w_{Y_4|X_2,Y_3}(a|x_2,y_3)\right),\\
A[2,b] &=& \min_{y_2y_3\in D(1)}\left(A[1,y_2y_3] + w_{Y_4|X_2,Y_3}(b|x_2,y_3)\right).\\
\end{eqnarray*}
}

\noindent
Then we have $w_X(x_1,x_2) = \min_{y_4\in D(2)}A[2,y_4]$. 
\EXX
\end{example}

Finally, note that in a Bayesian network with a non-graded structure the inference algorithm given by the evaluation of the expression $w_X(x)$ has generally
a more complex bookkeeping structure for the computation of the expression~(\ref{e-grad}),
since it requires to resort on values of hidden variables with arbitrarily small semi-rank (Fig.~\ref{f-N5}).
The corresponding data structure (array) $A$ holding these values will be rather intricate and meander-shaped.

\end{document}